# News Verifiers Showdown: A Comparative Performance Evaluation of ChatGPT 3.5, ChatGPT 4.0, Bing AI, and Bard in News Fact-Checking

Kevin Matthe Caramancion
*Mathematics, Statistics, and Computer Science Department*
*University of Wisconsin–Stout*
Menomonie, Wisconsin, United States
caramancionk@uwstout.edu / www.kevincaramancion.com

*Abstract*—This study aimed to evaluate the proficiency of prominent Large Language Models (LLMs)—OpenAI's ChatGPT 3.5 and 4.0, Google's Bard/LaMDA, and Microsoft's Bing AI—in discerning the truthfulness of news items using black box testing. A total of 100 fact-checked news items, all sourced from independent fact-checking agencies, were presented to each of these LLMs under controlled conditions. Their responses were classified into one of three categories: True, False, and Partially True/False. The effectiveness of the LLMs was gauged based on the accuracy of their classifications against the verified facts provided by the independent agencies. The results showed a moderate proficiency across all models, with an average score of 65.25 out of 100. Among the models, OpenAI's GPT-4.0 stood out with a score of 71, suggesting an edge in newer LLMs' abilities to differentiate fact from deception. However, when juxtaposed against the performance of human fact-checkers, the AI models, despite showing promise, lag in comprehending the subtleties and contexts inherent in news information. The findings highlight the potential of AI in the domain of fact-checking while underscoring the continued importance of human cognitive skills and the necessity for persistent advancements in AI capabilities. Finally, the experimental data produced from the simulation of this work is openly available on Kaggle.

## I. Introduction

One promising application of large language models (LLMs) is fact-checking news information. As an emerging authoritative technology across multiple domains, LLMs will undisputedly attract a mass of users seeking information confirmation. Preliminary findings on the accuracy of these models, however, are conflicting; That is, performance data yields suggest that on small sample sizes, they perform virtually in perfection [1]. On the other hand, hallucinations of these models are known to exist equally on both simple and complicated question prompts, showcasing their unpredictability [2].

The currently dominant LLMs in the US markets are Open AI's ChatGPT (based on Generative Pre-training Transformer or "*GPT*" 3.5 and 4.0 models), Google's Bard/*LaMDA*—short for "Language Model for Dialogue Applications," and Microsoft's Bing AI or also known as the *Sydney* anchored on Prometheus AI model. One of the striking distinctions in these LLMs is the training data used to arrive at the generative response provided by each model. There is no unison or common design between these LLMs; as such, generalizations on their performance in the very endeavor of *fake news* detection may be, in itself, *misleading*.

For instance, one striking limitation (or feature) in ChatGPT's legacy models up to 3.5, including the standard 4.0, is its lack of input inclusion of the web data after its knowledge cutoff date [3]. As of this writing, a *beta* version that allows it to browse the web for additional input is available as a supplement to the GPT 4.0 model (i.e., the *browsing* mode) [3]. The two chatbots, Bard and Sydney, are both without such restrictions in accessing the web in their generative responses. Fascinatingly though, a quick inspection of their responses' formatting, such as length, topic saturation, correctness, etc., are visibly non-similar. On top of these, ChatGPT's premium version is not free and imposes a cap on response count per time period.

Multiple metrics for comparative evaluations between the existing and future LLMs can be considered to decide which one performs most optimally. As a constitutive reference, this paper's sole focus, however, is on the dimension of misinformation and disinformation detection *accuracy* [4]. An investigative style of testing will be applied to the three prevailing LLMs discussed above, and the findings are all based on the experiment performed in this work. No existing assumptions nor performance expectations are considered prior to the inception of this project.

This paper will straightforwardly test the three LLMs', ChatGPT, Bard, and Bing AI,'s accuracy in a simulation where they will each be discerning facts from deceptive news information using a common prompt. This paper anchors on authoritative, independent fact-checking agencies in the United States, such as PolitiFact and Snopes; We then cross reference each model's response to that of the currently established "truth" as vouched for by these independent fact-checkers. We then analyze and present the performance evaluation of these models solely on one metric: the accuracy in correctly classifying news headline articles according to their integrity.

The following are the two main research questions we seek to substantiate. This paper will center on illuminating the two inquiries below:

RQ1: How accurately do leading large language models, specifically OpenAI's ChatGPT, Google's Bard/LaMDA, and Microsoft's Bing AI, discern facts from deceptive news information in a controlled simulation?

RQ 2: How does their performance vary in relation to established fact-checking agencies such as PolitiFact and Snopes?

The literary contribution of this paper is the addendum it confers to the continually and rapidly growing applications of LLMs. This paper's main topical application, *fake news,* is inherently grounded in several interdisciplinary domains, including but not limited to media studies, artificial intelligence, and education [5]. Additionally, this paper utilizes fundamental principles and constructs in its method, such as psychological experimental designs and mathematical modeling—touching on both qualitative and quantitative perspectives on its analysis and interpretations.

The practical contribution of this paper is the insight it provides on the potential role of LLMs are *the solutio*n for combatting cyber deceptions—misinformation and disinformation—and their advanced representations in the form of deepfakes and other AI-powered content. Dependency on human skills or information and media literacy alone is insufficient to destroy such falsehoods. An even more effective way of refuting these deceptions requires supplementary technological interventions as dictated in the disinformation ecosystem model proposed and theorized by [6] and proven in [7].

Finally, this paper is formatted as follows: The subsequent sections provide the necessary briefings on the following contexts to holistically provide the readers with the necessary background to understand the simulation process utilized in this paper (a) the fundamentals of *fake news* tests/experiments, (b) mis/disinformation cyber risk modeling, and (c) prior works related to LLMs involving fake news. Afterward, the third chapter thoroughly explains the steps performed in the simulation so readers may re-create this experiment if desired. As an addendum, limitations inherent to this paper's simulation are also contained herein. The presentation of findings is placed in a separate chapter following the methods and materials section presenting the simulation results. A discussion section that analyzes the results in-depth will be the penultimate chapter of this paper. Finally, this paper concludes with the recommendations and directions for future iterations of this research area.

## II. Literary Background

### A. Psychometry of Fake News Assessment Devices

In the context of misinformation and disinformation, *vulnerability* refers to the likelihood of an *agent*, typically human, falling prey to believing that disseminated false news information is correct [8]. The primal motivation in experimentations seeking to gauge the capability of *agents* to thwart deceptive social media content is rooted in the fact that there are very few to virtually no publicly available datasets to exhibit such vulnerability [9]. Furthermore, the social networking giant Facebook banned researchers involved in similar and related works, suggesting that such activities go against the terms and policies of the company, which critics have argued a move made to preserve the public image and branding of the platform [10]. As such, the experiments attempt to mimic an actual social platform by presenting non-synthetic news items, that is, actual content found on social media sites [11].

The actual simulation consists of both legitimate and misleading news items. The format of the experimentations typically varies; More controlled studies hold in-person tests where printouts of news headlines are given to the participants, and they then decide the legitimacy of the items. Stricter study designs totally prohibit access to electronic devices and the web to confirm the information presented in the test items. More modern variations of the study design transitioned to electronic testing, with most deployed over the web to distribute the experiment survey across a wider group of participants with diverse backgrounds [9].

### B. Mis/Disinformation Cyber Risk Modeling

The performance simulations involving the agents yield two performance data: First is the accuracy of their performance, that is, the correct number of detections (as count data) among the total count of presented items. Secondly, the time it takes for the agents to finish their assessments is typically measured in seconds. In the succeeding analyses, these garnered performance data are depicted as dependent variables on a wide breadth of a set of possible predictors—all in the name of attributing the possible correlative factors that may have caused performance gains or losses in the resulting experimental data [9].

The tested areas include demographic and socioeconomic factors, including the agent's age [12], sex assigned at birth [12], household income [9], native language [13], religion [14], veteran status [15], etc. Other areas include psychological perceptions [16] and political positions [17] as possible predictors, although these proposals currently yield non-significant outcomes. In the information sciences, studies focus on the test items and not directly on the agents; that is, some examples of the proposed predictors are the modality or textuality of news content [18], contextual clues that come with the news items [11], other metadata such as the time of circulation and exposure of news items to social media users [19]. On the other hand, predictors in behavioral sciences include the reporting behavior of users when doubting social media content [20] and the preferred device type when browsing a social media website [21]. Finally, arguments establishing that the way an information environment is designed itself, including but not limited to social media policies and terms and use of agreements, can either promote or demote mis/disinformation content [22][23].

*C. Large Language Models (LLMs) in Fake News Management*

The philosophical rooting of the proposal that LLMs be used to combat misinformation and disinformation [1] is the argument that there are forms of *fake news* that are advanced for the typical social media users to recognize easily and quickly, regardless of their information literacy skills. These deceptions are versed to be inherently cybersecurity threats [6][7] as they are carefully engineered for the sole purpose of swaying public opinions and causing societal polarization. An important but practical study likened such to distributed denial of service (DDoS) attacks as they typically come in mass and waves, especially at times of interest such as elections and public emergencies where civil unrests are highly likely to affect the national security of a nation [22].

Naturally, since cyber deceptions powered by advanced technologies are difficult for humans to combat with skills alone, the proposed ecosystem of [1] promotes the idea that the same technologies be used to combat such powerful mistruths. The use of machine and deep learning, for instance, to detect *deepfakes,* is currently a rich and thriving stream of inquiry in this stream. Additionally, disinformation-as-service disseminated by bots are challenging for independent fact-checkers alone to combat, and thus, LLMs come in the role of balancing the scale of such information ecosystem.

*D. The Literary Gap: Comparative Evaluation of the Performance of LLMs in Mis/Disinformation Detection*

Given these underpinnings, the following corollaries are evident:

a. That the existing proposed *disinformation ecosystem model* of [6] to use technologically powered solutions to technologically powered deceptions is not without its merit and appears to be very compelling.

b. That an existing study initiated by [1] showcasing and gauging the ability of an *LLM as an agent* instead of actual humans exists and that—

c. Although the said study presented promising results and outcomes—the study design and sample size are premature and have been called on to be improved upon in future studies.

d. This paper answers the open call of the previous paper and fulfills the currently existing limitation and gaps. We improve upon the study design of [] by making a more comprehensive and controlled experimental simulation, adding more LLMs, and expanding the count of included test items.

III. EXPERIMENTAL DESIGN & SIMULATIONS

*A. Overview*

This chapter details the methodological approach and materials used in conducting this study. It outlines the selection of large language models (LLMs), data collection procedures, simulation setup, metrics for evaluation, data analysis process, and the limitations of the study design. In addition, it also addresses reproducibility and ethical considerations. Note that the experimental data produced from this simulation is openly available on Kaggle [25].

*B. Selection of the Large Language Models (LLMs)*

1) *Brief Description of Selected LLMs*

**OpenAI's ChatGPT 3.5 and 4.0:** This LLM, trained by OpenAI, is built on Generative Pre-training Transformer (GPT) technology and provides generative responses. Its unique characteristic is its knowledge cutoff, which limits the inclusion of web data beyond a certain date in its training.

The primary difference between GPT-3.5 and GPT-4 lies in their capacity for understanding and generating text. As a newer iteration, GPT-4 is more powerful, trained on a broader dataset, and has improved capabilities in generating coherent and contextually relevant text. This enhanced capability is due to its larger number of parameters, enabling it to learn more complex patterns in language. Additionally, GPT-4 may exhibit better performance in tasks that require a deep understanding of context and abstraction. However, both versions have a similar knowledge cutoff, meaning they can't generate responses based on real-world events or information beyond their respective training periods.

**Google's Bard/LaMDA:** Bard/LaMDA is Google's LLM designed for dialogue applications. Unlike ChatGPT, it doesn't have restrictions regarding accessing the web for generating responses.

**Microsoft's Bing AI or Sydney:** Anchored on the Prometheus AI model, Bing AI or Sydney is Microsoft's LLM offering. It doesn't have web data access restrictions, making it different from ChatGPT.

*C. Data Collection*

1) *Collection of News Headlines*

   *a) Sourcing from News Outlets*

We collected fact-checked and verified content from independent fact-checking agencies and cross-referenced them. As a form of additional validation, we verified and checked the posting and content modification dates of the items to confirm that no updates were added to their legitimacy. The reason for this is that a news item may change its truth value over time when refuted (i.e., True to False) or when proven otherwise (i.e., False to True).

   *b) Criteria for Inclusion*

All the news items presented to the LLMs are up to September 2021 to level the playing field for ChatGPT since its knowledge cutoff date is only until the specified date. Furthermore, news items that are unverified by independent fact-checking agencies, albeit posted by any media, are not included in the pool of news items presented. This control is put in place to limit the framing bias that may be yielded in the simulation by media, including state-owned, publicly owned, or even private, for-profit organizations; We only consider the independent fact-checkers as the sole source of the truth for this experiment.

*c) Procedure for Classification as True or False*

The attached legitimacy for the test items is threefold—(1) True, (2) False, and (3) Partially True or False. The basis of this simulation, unlike its paper predecessor, doesn't rely solely on binary classification/option responses. We observed that the fact-checked content of the third-party agencies typically falls into four categories—(1) True, (2) False, (3) Partially True, or (4) Partially False. In this experiment, we combine the latter two on the grounds that the original categories of "Partially True" and "Partially False" can be ambiguous and potentially confusing. We strive to eliminate this ambiguity and make it clear that these items contain a mix of true and false information. Practically, in many cases, it may be difficult to distinguish between "Partially True" and "Partially False" items.

D. *Simulation Setup*

 1) *Testing Environment*

  *a) Details of the Control Environment*

The testing environment was carefully designed to ensure a fair and controlled evaluation of the Language Learning Models (LLMs). The LLMs were run on identical hardware configurations to avoid any performance discrepancies due to hardware differences. The software environment was also standardized, with all LLMs running on the same operating system and using the same versions of necessary libraries and dependencies. This ensured that any differences in the LLMs' performance could be attributed to the models rather than external factors.

  *b) Control Measures to Prevent Uncontrolled Variables*

Several control measures were implemented to prevent uncontrolled variables from influencing the results. First, the LLMs were all tested under the same conditions, at the same time of day, to avoid any potential effects of network traffic or server load. Second, the same set of news headlines was presented to each LLM, ensuring that all models were evaluated on the same tasks. Finally, any updates or modifications to the LLMs were prohibited during the testing period to maintain consistency.

 2) *Testing Procedure*

  *a) Prompts Used for Testing LLMs*

The LLMs were tested using a set of prompts derived from the collected news headlines. Each prompt was designed to elicit a response that could be classified as true, false, or partially true/false. The prompts were presented to the LLMs in a random order to avoid any potential order effects.

  *b) Timeline of Simulation*

The simulation was conducted over a period of one month, from mid-May to mid-June 2023. This timeline was chosen to allow sufficient time for the LLMs to process the prompts and for the researchers to analyze the results. Each LLM was tested individually, with a one-week interval between each testing session to allow for any necessary adjustments or troubleshooting.

E. *Metric for Evaluation*

 1) *Definition of Accuracy*

Calculation of Detection Accuracy: Accuracy was defined as the proportion of correct classifications (true, false, partial) by the LLMs compared to the established truth from the fact-checking agencies.

We did not include the second metric, the time it takes for an LLM to produce the response, and it is part of a different project. As a possible confounding factor, the time it takes for an LLM to produce a response could be influenced by the complexity of the input, the processing power of the computer, and the specific implementation of the LLM. By not including this metric, we strive to avoid the need to control for these potentially confounding factors.

F. *Limitations of the Study Design*

1. To eliminate the classification ambiguity of the LLMs, we forcibly asked in the prompts on each LLM to choose between three choices (True, False, Partially True/False) for the elicitation of an even more explicit response.

2. Additionally, our evaluation metric focuses solely on the accuracy of the LLMs' classifications. This means that this study does not consider other important aspects of performance, such as the speed of response or the quality of the generated text.

3. Finally, we repeatedly highlighted the independent fact-checking agencies as our source; however, these agencies are not infallible, and there is a potential for misclassification. If an agency incorrectly classifies a news item, this could unfairly penalize the LLMs in our evaluation.

G. *Ethical Considerations*

1. Ensuring the protection of user data was paramount during the simulation, adhering strictly to privacy policies and confidentiality principles.

2. The study was conducted with a keen emphasis on responsible AI principles, such as fairness, accountability, transparency, and ethics in AI usage.

## IV. FINDINGS & RESULTS

Note: The performance data of the LLMs are openly available and can be accessed on Kaggle. The succeeding analyses are based on [25]

A. *Descriptive Statistics*

 1) *Average Accuracy Rates*

TABLE 1: SUMMARY OF LLMS' PERFORMANCE

| | Descriptive Statistics | | | | | | | | |
|---|---|---|---|---|---|---|---|---|---|
| | N Statistic | Range Statistic | Minimum Statistic | Maximum Statistic | Sum Statistic | Mean Statistic | Mean Std. Error | Std. Deviation Statistic | Variance Statistic |
| Correct_Detections | 4 | 9.00 | 62.00 | 71.00 | 261.00 | 65.2500 | 1.97379 | 3.94757 | 15.583 |
| Incorrect_Response | 4 | 9.00 | 29.00 | 38.00 | 139.00 | 34.7500 | 1.97379 | 3.94757 | 15.583 |
| Valid N (listwise) | 4 | | | | | | | | |

B. *Comparative Analysis*

 1) *Differences Among LLMs*

TABLE 2: INDIVIDUAL PERFORMANCE OF EACH LLMS

| LLM_name | | Correct_Detections | Incorrect_Response |
|---|---|---|---|
| Bard | Mean | 64.0000 | 36.0000 |
| | Median | 64.0000 | 36.0000 |
| | Range | .00 | .00 |
| Bing | Mean | 64.0000 | 36.0000 |
| | Median | 64.0000 | 36.0000 |
| | Range | .00 | .00 |
| GPT-3.5 | Mean | 62.0000 | 38.0000 |
| | Median | 62.0000 | 38.0000 |
| | Range | .00 | .00 |
| GPT-4.0 | Mean | 71.0000 | 29.0000 |
| | Median | 71.0000 | 29.0000 |
| | Range | .00 | .00 |
| Total | Mean | 65.2500 | 34.7500 |
| | Median | 64.0000 | 36.0000 |
| | Range | 9.00 | 9.00 |

*2) Performance Plot of the LLMs*

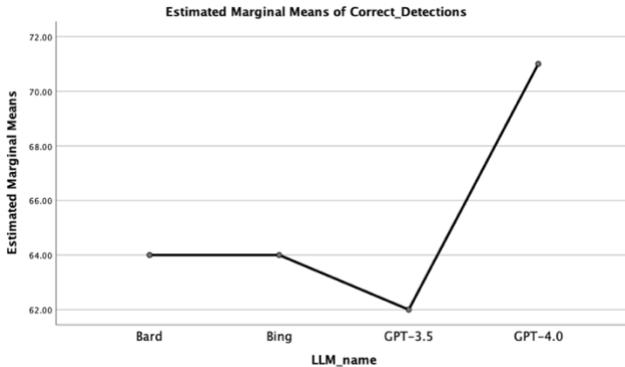

## V. ANALYSES AND INTERPRETATIONS

*Discussion 1: LLMs, On Average, Scored 65% of the Total Points Possible*

Response to the First Research Question (RQ1): In this study comparing the proficiency of leading large language models (LLMs) in discerning facts from deceptive news information, OpenAI's GPT-4.0 outperformed its predecessor GPT-3.5, Google's Bard/LaMDA, and Microsoft's Bing AI in a controlled simulation. The average score across all tested models was 65.25 out of a possible 100 points, indicating a moderate ability to discern factual from deceptive information correctly. GPT-4.0 scored the highest with 71 points, suggesting that newer iterations of LLMs may have improved capabilities in this critical area. However, the difference in performance among the models was not significant. Both Google's Bard/LaMDA and Microsoft's Bing AI scored 64, slightly below the average but still comparable to GPT-3.5, which scored 62.

While all these models demonstrate competency in discerning facts from deceptive news, there is evidently considerable room for improvement. These findings underscore the importance of ongoing research and refinement in the field of large language models, especially concerning the crucial task of discerning facts from mis/disinformation. Furthermore, this research prompts further investigation into how these models can be better trained to improve their accuracy in differentiating factual from deceptive information.

*Discussion 2: LLMs Show Promise but Human Fact-Checkers Still More Reliable*

Response to the Second Research Question (RQ2): The performance of the large language models (LLMs) like OpenAI's GPT-3.5 and GPT-4.0, Google's Bard/LaMDA, and Microsoft's Bing AI in discerning facts from deceptive news information was found to be moderately successful with an average score of 65.25 out of 100. When compared with established fact-checking agencies such as PolitiFact and Snopes, these AI models showed considerable promise but still fell short of the rigorous analysis carried out by human fact-checkers.

GPT-4.0, the highest-scoring model with 71 points, demonstrated progress in this area, but it is important to note that independent fact-checking agencies often delve deeper into the context and nuances of claims, corroborating information from multiple sources. Their processes typically involve expert knowledge, meticulous research, and the human ability to discern subtleties and contexts that AI currently struggles with. Both Bard/LaMDA and Bing AI scored 64, whereas GPT-3.5 scored slightly lower at 62, indicating that while these models have a certain level of proficiency in fact-checking, they are not yet at par with human-run agencies.

This comparison highlights the potential and limitations of AI in the realm of fact-checking. It illustrates the necessity for ongoing research and development in improving the ability of AI models to understand and evaluate nuanced information. However, it also emphasizes that, as of now, human-led fact-checking agencies continue to provide a more reliable check against deceptive news information.

## VI. CONCLUSION & FUTURE WORKS

In the ceaseless war against disinformation and misinformation, this research underscores the compelling potential and the profound limitations of Large Language Models (LLMs) as potent allies. Our examination of three leading LLMs–OpenAI's ChatGPT 3.5 and 4.0, Google's Bard/LaMDA, and Microsoft's Bing AI–highlighted their moderate proficiency in discerning fact from fabrication, with an average score of 65.25 out of a possible 100. Yet, even as the newest iteration, GPT-4.0, led the pack with 71 points, we are reminded that these AI models still lag behind the rigorous, context-aware analysis executed by human fact-checkers in established agencies such as PolitiFact and Snopes.

Nonetheless, we should not diminish the progress these AI models represent. The improvements exhibited by GPT-4.0 point to a future where AI models will become increasingly adept at navigating the complex landscape of news and information. However, this potential future does not absolve us of the pressing present. The insidious threat of cyber deceptions remains rampant, and as we look to AI for aid, we must continue to foster human skills and literacy, nurturing a complementary symbiosis between man and machine.

The juxtaposition of AI capabilities and human expertise offers a poignant reflection of our times. AI's emergence as a powerful tool in the fight against misinformation marks a turning point in our journey, yet it also highlights the irreplaceable value of human cognition, judgment, and emotional intelligence. The growth of AI, therefore, should not be perceived as a journey towards human redundancy but rather as an opportunity for harmonious collaboration.

In an era increasingly shaped by information integrity, our survival and success hinge not only on technological innovation but also on our ability to integrate these advancements with the cognitive capacities that make us uniquely human. It is in this synergy that we can foster a robust defense against the relentless onslaught of misinformation, ensuring a future where truth triumphs over deception.

REFERENCES

[1] Caramancion, K. M. (2023, June). Harnessing the Power of ChatGPT to Decimate Mis/Disinformation: Using ChatGPT for Fake News Detection. In *2023 IEEE World AI IoT Congress (AIIoT)*. IEEE.

[2] Alkaissi, H., & McFarlane, S. I. (2023). Artificial hallucinations in ChatGPT: implications in scientific writing. *Cureus*, *15*(2).

[3] OpenAI. (2023). Knowledge cutoff date of September 2021. *Documentation*. Retrieved via openai.platform.com

[4] Caramancion, K. M. (2022, June). Using Timer Data to Conjunct Self-Reported Measures in Quantifying Deception. In *2022 IEEE World AI IoT Congress (AIIoT)* (pp. 065-070). IEEE.

[5] Caramancion, K. M. (2023, July). An Interdisciplinary Perspective on Mis/Disinformation Control. In *2023 International Conference on Electrical, Computer, Communications and Mechatronics Engineering (ICECCME)* (pp. 1-6). IEEE.

[6] Caramancion, K. M. (2020, March). An Exploration of Disinformation as a Cybersecurity Threat. In *2020 3rd International Conference on Information and Computer Technologies (ICICT)* (pp. 440-444). IEEE.

[7] Caramancion, K. M., Li, Y., Dubois, E., & Jung, E. S. (2022, April). The Missing Case of Disinformation from the Cybersecurity Risk Continuum: A Comparative Assessment of Disinformation with Other Cyber Threats. *Data*, 7(4), 49.

[8] Caramancion, K. M. (2022, November). Quantification of Infographic Intervention Effect on Mis/Disinformation Vulnerability. In *2022 IEEE 17th International Conference on Computer Sciences and Information Technologies (CSIT)* (pp. 297-300). IEEE.

[9] Caramancion, K. M. (2022). *An Interdisciplinary Assessment of the Prophylactic Educational Treatments to Misinformation and Disinformation*. State University of New York at Albany.

[10] Bond, S. (2021, August). NYU Researchers Were Studying Disinformation On Facebook. The Company Cut Them Off. *Technology*. NPR. Retrieved via npr.org

[11] Caramancion, K. M. (2020, September). Understanding the Impact of Contextual Clues in Misinformation Detection. In *2020 IEEE International IOT, Electronics and Mechatronics Conference (IEMTRONICS)* (pp. 1-6). IEEE.

[12] Caramancion, K. M. (2021, April). The Demographic Profile Most at Risk of being Disinformed. In *2021 IEEE International IOT, Electronics and Mechatronics Conference (IEMTRONICS)* (pp. 1-7). IEEE.

[13] Caramancion, K. M. (2022, January). The Role of User's Native Language in Mis/Disinformation Detection: The Case of English. In *2022 IEEE 12th Annual Computing and Communication Workshop and Conference (CCWC)* (pp. 0260-0265). IEEE.

[14] Caramancion, K. M. (2023, March). The Link Between a User's Religion and Mis/Disinformation Vulnerability. In *2023 IEEE 13th Annual Computing and Communication Workshop and Conference (CCWC)* (pp. 0129-0133). IEEE.

[15] Caramancion, K. M. (2022, October). Veteran Status as a Potent Determinant of Misinformation and Disinformation Cyber Risk. In *2022 IEEE 13th Annual Ubiquitous Computing, Electronics and Mobile Communication Conference (UEMCON)* (pp. 0040-0044). IEEE.

[16] Caramancion, K. M. (2021, October). The Role of Subject Confidence and Historical Deception in Mis/Disinformation Vulnerability. In *2021 IEEE 12th Annual Information Technology, Electronics and Mobile Communication Conference (IEMCON)* (pp. 0541-0546). IEEE.

[17] Caramancion, K. M. (2021, May). Understanding the Association of Personal Outlook in Free Speech Regulation and the Risk of being Mis/Disinformed. In *2021 IEEE World AI IoT Congress (AIIoT)* (pp. 0092-0097). IEEE.

[18] Caramancion, K. M. (2021, October). Textual vs Visual Fake News: A Deception Showdown. In *2021 IEEE International Conference on Cloud Computing in Emerging Markets (CCEM)*. IEEE.

[19] Caramancion, K. M. (2021, September). The Relation Between Time of the Day and Misinformation Vulnerability: A Multivariate Approach. In *2021 IEEE 16th International Conference on Computer Sciences and Information Technologies (CSIT)* (Vol. 1, pp. 150-153). IEEE.

[20] Caramancion, K. M. (2021, December). The Relation of Online Behavioral Response to Fake News Exposure and Detection Accuracy. In *2021 IEEE 12th Annual Ubiquitous Computing, Electronics and Mobile Communication Conference (UEMCON)* (pp. 0097-0102). IEEE.

[21] Caramancion, K. M. (2022, October). Preference of Device on Social Media Browsing as a Predictor of Deception: The Link Between UX and Mis/Disinformation Vulnerability. In *2022 IEEE 13th Annual Information Technology, Electronics and Mobile Communication Conference (IEMCON)* (pp. 0583-0588). IEEE.

[22] Caramancion, K. M. (2021, November). The Role of Information Organization and Knowledge Structuring in Combatting Misinformation: A Literary Analysis. In *International Conference on Computational Data and Social Networks* (pp. 319-329). Springer, Cham.

[23] Caramancion, K. M. (2022, June). An Exploration of Mis/Disinformation in Audio Format Disseminated in Podcasts: Case Study of Spotify. In *2022 IEEE International IOT, Electronics and Mechatronics Conference (IEMTRONICS)* (pp. 1-6). IEEE.

[24] Caramancion, K. M. (2022, June). Same Form, Different Payloads: A Comparative Vector Assessment of DDoS and Disinformation Attacks. In *2022 IEEE International IOT, Electronics and Mechatronics Conference (IEMTRONICS)* (pp. 1-6). IEEE.

[25] Kevin Matthe Caramancion. (2023). <i>LLM Comparative Performance Fake News Detection_v1</i> [Data set]. Kaggle. https://doi.org/10.34740/KAGGLE/DSV/5959587